\documentclass[runningheads]{llncs}
\usepackage{graphicx}
\usepackage[OT2,T1]{fontenc}
\usepackage{amsmath}
\usepackage{amssymb}
\usepackage{booktabs}
\usepackage{multirow}
\usepackage{color, colortbl}
\definecolor{LightCyan}{rgb}{0.88,1,1}
\usepackage{comment}
\usepackage{algorithm}
\usepackage{algpseudocode}
\begin{document}
%
\title{Slovo: Russian Sign Language Dataset}
%

\author{Kapitanov Alexander \and
Kvanchiani Karina \and Nagaev Alexander \and Petrova Elizaveta}

\authorrunning{Kapitanov A. et al.}
\institute{SaluteDevices, Russia \\
\email{\{aakapitanov, kskvanchiani, aonagaev, emikhaylpetrova\}@sberbank.ru}}

\maketitle              
\begin{abstract}
One of the main challenges of the sign language recognition task is the difficulty of collecting a suitable dataset due to the gap between hard-of-hearing and hearing societies. In addition, the sign language in each country differs significantly, which obliges the creation of new data for each of them. This paper presents the Russian Sign Language (RSL) video dataset Slovo, produced using crowdsourcing platforms. The dataset contains 20,000 FullHD recordings, divided into 1,000 classes of isolated RSL gestures received by 194 signers. We also provide the entire dataset creation pipeline, from data collection to video annotation, with the following demo application. Several neural networks are trained and evaluated on the Slovo to demonstrate its teaching ability. Proposed data and pre-trained models are publicly available\footnote{\url{https://github.com/hukenovs/slovo}}\footnote{\url{https://gitlab.aicloud.sbercloud.ru/rndcv/slovo}}.

\keywords{Sign Language \and Video Dataset \and Gesture Recognition \and Data Creating Pipeline \and Human Computer Interaction.}
\end{abstract}

\section{Introduction}
While the contemporary world is developing rapidly with the advent of high-end technologies, some parts of society are out of their scope. One such part is the hard-of-hearing community, which still struggles in many situations and can be misunderstood in some extreme cases. For example, some hospitals still do not have a sign language interpreter on staff. Therefore the interaction between hard-of-hearing people and healthcare providers is complex, which prevents timely assistance. A similar problem exists in structures such as banks, government institutions, airports, public places, and others, significantly complicating their everyday life. Moreover, many consequences of deaf as social isolation, an education gap with the hearing population, and difficulties in finding employment, also negatively affect the life of this community. Sign Language Recognition (SLR) systems have the potential to simplify these processes by, for example, developing a sign language learning app~\cite{asl-signs} or embedding a feature in video conferencing apps. Also, such technology can accomplish more transparent communication between people with different hearing and speaking abilities and be integrated into human-computer interaction systems~\cite{kenshimov2021sign}, allowing hard-of-hearing individuals access to information and services easier and helping to overcome barriers in education~\cite{school} and employment.

\begin{figure}
  \centering
  \includegraphics[width=\linewidth] {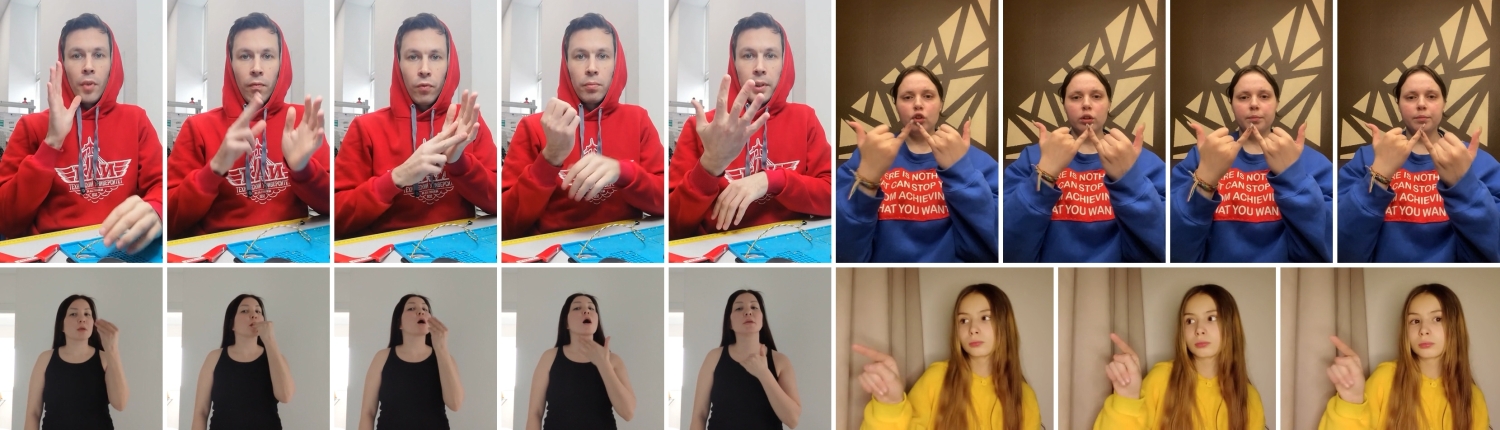}
  \caption{RSL signs "at eight fifteen" (left top), "appetite" (left bottom), "yellow" (right top), and "this" (right bottom).}
  \label{fig: main}
\end{figure} 

SLR is a field of study that should accurately convert sign language gestures from video footage into textual representation. This task is indispensable but daunting due to the tangle and rapid nature of sign language, which entails intricate hand gestures, body postures, and facial expressions. The complexity of data collection is the major problem of SLR due to a gap between hard-of-hearing and hearing communities. Adding to this the need for a different sign language for each country and significant language differences within even one country, Russian SLR system developers face the challenge of data absence. Furthermore, existing RSL datasets have few samples or must be sufficiently diverse across subjects, which is necessary to train a robust model. This paper presents two main contributions to simplify the solution of sign language recognition:

\begin{itemize}
\item We provide a pipeline for creating a video dataset consisting of three main steps: video collection, validation, and time interval annotation. Crowdsourcing platforms were used throughout the pipeline to increase the number of signers and improve the dataset's quality. We apply some exam tasks to signers for the most correctly executed gestures and add a quality check to the validation step to extract inappropriate videos. In the third step, all videos were marked by the start and end time of the gesture.
\item We release the Russian Sign Language dataset, Slovo, which can become the basis for this area. It consists of 20,000 FullHD videos from 194 signers and is divided into 1,000 classes of glosses from the RSL without words shown by dactyl (finger-spelling) or compound gestures. Figure~\ref{fig: main} shows the examples of gestures in our dataset. All signers recorded videos mostly in their homes or office in front of a laptop or smartphone camera. Each video, whose length varies from 1 second to 4, was cropped using two timestamps (start and end), contributing to the production of the "no event" class. Added class corresponds to the video's parts, where the signer is preparing to perform the gesture or has already performed it.
\end{itemize}


\section{Related Work}

This paper solely focuses on comparing our findings with the Russian Sign Language. It would be inappropriate to compare RSL datasets with SL datasets in foreign languages as they all differ significantly in structure. However, we provide an overview of datasets in other languages to show the specifics and features of this domain. Table~\ref{tabl:related} encompasses the prevalent datasets having significant data volumes. Since the collection step is the main problem of SL dataset creation and one of our contributions, we describe the main ways to collect it.
\vspace{-0.3cm}
\begin{table*}
\caption{The main characteristics of the reviewed SL datasets. Datasets are divided into two categories: isolated datasets, containing individual gestures, and continuous datasets, containing sentences in sign language. In addition, we added a column named "Method" with the collection way information since it says a lot about the videos. "Manually" means that the dataset's authors recorded the video and could influence the scenes, gestures, and recording. "Download" means that the authors downloaded videos from third-party resources that sometimes guarantee the high accuracy of the gestures shown. "Crowd." means that the authors used crowdsourcing platforms to collect videos. "Subset" means that RWTH-BOSTON-400 was created from another dataset BU ASL~\cite{Vogler2012ANW}.}
\label{tabl:related}
\centering
\scalebox{0.8}{
\begin{tabular}{|l|c|c|c|c|c|c|}
\hline
Dataset & Classes & Videos & Signers & Resolution & Language & Method \\
\hline\hline
\multicolumn{7}{|c|}{Continuous}\\
\hline
GSLC, 2007~\cite{efthimiou2007gslc} & 20 & 840 & 6 & 848 × 480 & Greek & manually\\
RWTH-BOSTON-400, 2008~\cite{dreuw2008benchmark} & 400 & 843 & 5 & 648 × 484 & American & subset\\
RWTH-PHOENIX-Weather, 2015~\cite{rwth-phoenix} & 1,066 & 8,257 & 9 & 210 × 260 & Germany & download\\
RSL, 2021~\cite{grif2021raspoznavanie} & 1,000 & 35,000 & 5 & FullHD & Russian & manually\\
\hline
\multicolumn{7}{|c|}{Isolated}\\
\hline
LSE-Sign, 2016~\cite{lse} & 2,400 & 2,400 & 2 & FullHD & Spanish & manually\\
LSA64, 2016~\cite{lsa} & 64 & 3,200 & 10 & FullHD & Argentinian & manually\\
MS-ASL, 2018~\cite{msasl} & 1,000 & 25,513 & 222 & varying & American & download\\
WLASL2000, 2020~\cite{wlasl} & 2,000 & 21,083 & 119 & varying & American & download\\
AUTSL, 2020~\cite{autosl} & 226 & 38,336 & 43 & 512 × 512 & Turkish & manually\\
TheRuSLan, 2020~\cite{kagirov2020theruslan} & 164 & 13 & 13 & FullHD & Russian & manually\\
K-RSL, 2020~\cite{imashev2020k} & 600 & 28,250 & 10 & FullHD & Kazakh-Russian & manually\\
FluentSigners-50, 2022~\cite{mukushev2022fluentsigners} & 278 & 43,250 & 50 & varying & Kazakh-Russian & crowd.\\
\rowcolor{LightCyan}
Slovo, 2023 (ours) & 1,000 & 20,000 & 194 & HD / FullHD & Russian  & crowd.\\
\hline
\end{tabular}}
\end{table*}
\vspace{-0.8cm}

\subsection{Sign Language Datasets in Russian Domain.}

There are four more widespread RSL datasets. The first, TheRuSLan~\cite{kagirov2020theruslan}, is composed of a total of 164 gestures, primarily related to the supermarket theme. A group of 13 signers was involved in the video collection, where each signer produced a unique recording with an average duration is 36 minutes. All signers come from different parts of the country, which generates variability within a class due to various dialects. The authors also proposed subtitles for each sample, which indicate the specific signs class. The second, FluentSigners-50~\cite{mukushev2022fluentsigners}, were created with the help of 6 natives, who chose frequently used signs, produced the templates for them, and wrote the instruction for signers. All signers came from different Kazakhstan regions, making the dataset a high degree of linguistics. Heterogeneity in the signer's age, skin color, clothes, variable background, and lighting make the dataset immensely diverse. The videos are in a total of 43 hours of labeled trimmed materials. The third, K-RSL~\cite{imashev2020k}, contains 4 subsets of phrases from a linguistic point of view: question-statement pairs, signs of emotion, emotional question-statement pairs, and phonologically similar signs. It was divided into 600 glosses with 28,250 examples in total. Ten signers recorded K-RSL, the first 5 are professional SL interpreters, and the other 5 are deaf native signers. The last, RSL~\cite{grif2021raspoznavanie}, consists of two sets of gestures obtained from an online dictionary. Each gesture was repeated by the signer at least 5 times. All signers are dressed in black suits against a clean background, which makes the dataset visually monotonous. All videos are marked with additional classes named "start gesture" and "end gesture"; suggestions include an additional class named "transition".

The TheRusLan and the FluentSigners-50 datasets are unsuitable for us due to the disuse of Kazakh-Russian Sign Language in Russia since the gestures are outdated. Besides, the TheRusLan dataset was created for only the supermarket domain and cannot be used for everyday life. Also, the two described datasets are not diverse in classes of signs and subjects. The RSL dataset can be used only in limited situations by us because we cannot influence the dataset's update by adding new sign classes. Furthermore, the RSL dataset was recorded by only 5 signers, which do not differ in clothing and background, complicating the training of a stable model. These reasons prompted us to create our dataset with 1,000 frequently used RSL signs received from 194 signers. We plan to extend it with new classes and increase the diversity of subjects.

\subsection{Others Sign Language Datasets.}

Since RSL differs from other sign languages, we describe only notable SL datasets, comparing them according to the main specific features of the domain. Many of reviewed datasets are not diverse in signers: RWTH-BOSTON-400~\cite{dreuw2008benchmark} were recorded by only 5 speakers, LSE-Sign~\cite{lse} -- by two sign language natives, LSA64~\cite{lsa} -- by 10 non-expert signers, and GSLC -- by 6 signers. Besides, LSE-Sign was recorded within one week to minimize the diversity of the signer's appearance. Others tried to make more heterogeneous datasets. MS-ASL~\cite{msasl} and WLASL2000~\cite{wlasl} are the most extensive publicly available ASL datasets, and their videos were produced by 222 and 119 signers, respectively. The AUTSL dataset~\cite{autosl} was recorded with 20 backgrounds, including indoor and outdoor scenes and with different angles. The reviewed datasets differ in the goal of creating and choosing the domain of signs. For example, The RWTH-PHOENIXWeather corpus~\cite{rwth-phoenix} contains SL recordings from the German TV station PHOENIX. The more typical variant to choose a sign basket is to collect it from the frequently daily-used signs like in AUTSL~\cite{autosl}. 

\subsection{Sign Language Dataset Collection.}
The main problem of dataset creation for SLR is video collection because it is challenging to find sign language experts. The need for diversity in signers makes this task even more problematic. The choice of sign basket is the other significant problem because natural and sign languages are highly different. We reviewed ways to collect sign language videos and divided this overview into three groups by collection methods for convenience. 

\textbf{Manually recorded videos.} One of the main ways to collect videos for sign language recognition is to produce them manually with a camera. Kagirov et al.~\cite{kagirov2020theruslan} used the MS Kinect 2.0 device to record video in 3D with a depth map to create the TheRusLan dataset. The Turkish Sign Language dataset, AUTSL~\cite{autosl}, was collected for real-life scenarios by the same camera. To make the model robust to scenes, 20 different backgrounds, including dynamic, various lighting conditions, from artificial light to sunlight, and different field-of-views were used to create AUTSL. The authors choose the frequently used signs; some are compound signs formed by simultaneously making two consecutive signs. The videos were performed by 43 different signers, where 60\% are students of the TSL course, 18\% are persons who know TSL (instructors and translators), 15\% are trained signers by the AUTSL dataset, and others related. The Argentinian Sign Language dataset, LSA64~\cite{lsa}, was recorded by a Sony HDR-CX240 camera in two different scene conditions: outdoor and indoor. The authors simplified the hand segmentation task with fluorescent-colored gloves. Signers wore black clothes against a white wall's backdrop for more accurate hand extraction. 

\textbf{Downloaded videos.} Another way to collect samples is to download from news or educational video sources. It has the advantage of correctly matching video and signs since sign language experts checked the content. For the WLASL2000 dataset, the authors~\cite{wlasl} chose multiple education SL websites as suitable video sources. They filtered samples by signs and leaved videos containing words only. Annotators of sign dialects were not native sign languages: they received training to understand SL specifics and, with a designed interface, compare signs from two videos displayed simultaneously. Samples for the MS-ASL dataset~\cite{msasl} were downloaded from video-sharing platforms to communicate and study ASL. Since videos are recorded and uploaded by ASL students and teachers, they differ by background, lighting, positioning, and dialect. Such platforms accompany the video with subtitles, which authors processed by OCR. Face detection and recognition are integral parts of sample preprocessing in cases where videos were taken from websites, and the authors included them in their dataset creation pipeline. 

\textbf{Kind of crowdsourcing.} Mukushev, Medet, et al.~\cite{mukushev2022fluentsigners} choose a more complicated but effective way to collect videos for SLR. Their dataset, FluentSigners-50, was created with six professional SL interpreters. They developed a sign basket including commonly useful phrases and sentences in the hard-of-hearing community. Other signers were invited by interpreters and use SL daily, and the subsequent distribution of signers by SL use was obtained: 32 deaf, 6 hard of hearing, 3 hearing SODA (Sibling of a Deaf Adult), and 9 hearing CODA (Child of Deaf Adults). They used instructions and templates from interpreters to repeat the KRSL sentences.

\section{Dataset Creation}
\begin{figure}
  \centering
  \includegraphics[width=\linewidth] {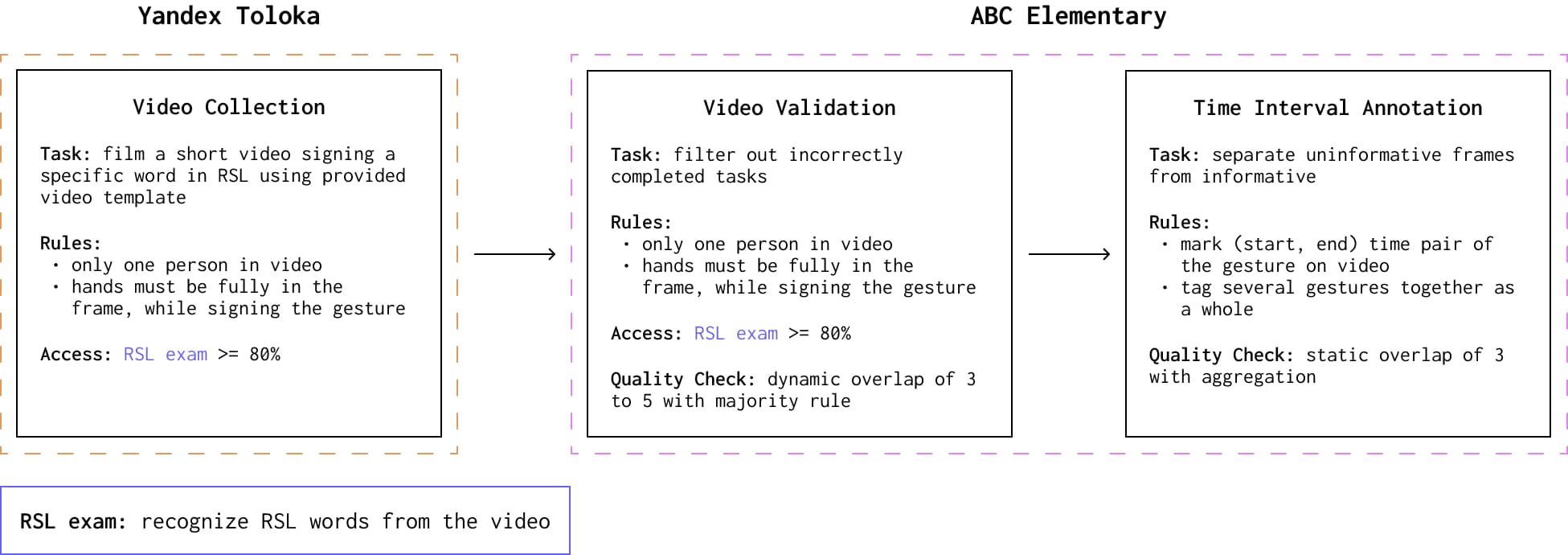}
  \caption{Crowdsourcing pipeline: collection, validation, and annotation. Each stage used its own rules, but the exam was the same.} 
  \label{fig: pipe}
\end{figure}

The following part provides details about our data collection pipeline. It consists of 3 main stages: (1) video collection, (2) video validation, and (3) and gesture time interval annotation. We used two crowdsourcing platforms: Yandex Toloka\footnote{\url{https://platform.toloka.ai/}} for data mining and ABC Elementary\footnote{\url{https://elementary.activebc.ru}} for the validation and the annotation so that different users are involved in recording and verifying videos. In addition, before each stage, crowdworkers must pass a mandatory RSL exam\footnote{The RSL exam aims to reveal the knowledge of the language, but it can be passed by language learners too.} with a score of at least 80\% before being granted access to assignments. These two nuances allow us to get a better and unbiased assessment of the correctness of the videos. Figure~\ref{fig: pipe} shows the main part of the dataset creation process.
\vspace{-0.3cm}

\vspace{-0.3cm}

\textbf{1. Video Collection.} The essential part of dataset creation started with designing a sign basket. We paid attention to choosing words frequently used in everyday life, and in the end, it turned out to be 1,000 glosses. We chose words related to commonly used topics such as food, animals, emotions, colors. After, we asked the crowdworkers from Yandex Toloka to record a short video of themselves singing a specific word in RSL based on the provided template. Video templates were taken from SpreadTheSign website\footnote{\url{https://www.spreadthesign.com/ru.ru/search/}}, a project of the European Sign Language Center association. Participants provided informed consent for data processing, ensuring compliance with legal requirements. No discrimination or bias was present in the dataset, promoting fairness and inclusivity. 

\textbf{2. Video Validation.} Correctly signing the gesture can be challenging for people not fluent in sign language, so we added the validation stage on the ABC Elementary platform in our dataset-creating pipeline. Workers were asked to check if the gesture was performed correctly. Each video was checked at least by three different workers. If they disagree, another marker participated in the validation of such a video, so up to 5 markers on the video could be repeated. If most workers mark the video as invalid, it is rejected; otherwise, it is accepted and passed to the next stage. After the validation, we left videos with a short edge of at least 720 pixels and converted them to a 30 fps rate.

\textbf{3. Time Interval Annotation.} Collected videos may contain uninformative frames at the beginning and the end of the video, where workers turn the camera on and off and prepare to show the gesture. Therefore, annotating the gesture's start and end time on the video is necessary. The crowdworkers from ABC Elementary were asked to indicate the time interval with a gesture. Since our dataset contains glosses and phrases, some videos may have several gestures. In this case, workers should tag them together as a full gloss. Each video was annotated by three different crowdworkers. 

\begin{figure}
  \centering
  \includegraphics[width=\linewidth] {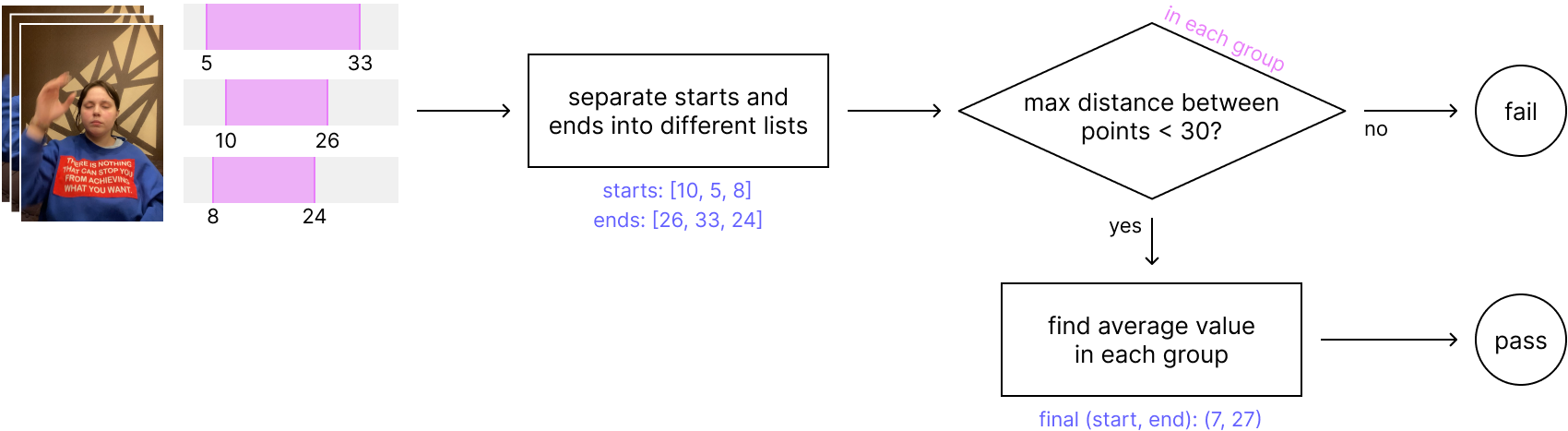}
  \caption{Time intervals aggregation pipeline. First, we split the beginning and end timestamps into different groups and then independently calculated distances between all points in each group. Then, if the maximum distance is less than 30 frames, we find the average value of each group and assume them to be the final pair (begin, end). Otherwise, video with such annotations was not taken into the dataset.}
  \label{fig: agg}
\end{figure}

Figure~\ref{fig: agg} shows the developed aggregation algorithm to get the average over the responses time interval. After cutting off the gestures, we had the cuts at the beginning and the end of the video where no gesture is shown, and we decided to use them as zero-class objects in training to predict the absence of action.

\vspace{-0.5cm}

\textbf{Dataset Post-processing.} While reviewing the collected videos, we noticed that some users show gestures significantly slower than others. This circumstance leads to inhomogeneous video length within the same class: the duration of the same gesture varied by more than five times, complicating the classification of such data. To make our dataset more homogenous, we decided to calculate the distribution of video lengths for each class and speed up those videos that are slower than the average value by more than 30 frames. As a result, 347 videos from 270 classes were sped up by an average of 1.7 times. In addition, we compared two variants -- with and without this processing -- and ensured that homogenous speed increases the accuracy of RSL recognition.

\section{Dataset Description}
\textbf{Dataset Content.} Our dataset is approximately 16 GB -- it contains 20,000 videos of 1,000 classes representing frequently used glosses and short phrases in Russian Sign Language, including alphabet and numerals. The dataset does not include fingerspelling words, i.e., words spelled letter by letter using dactylology. In addition, we expanded the Slovo by 400 extra samples of a special “no event” class where the subject is not signing any gestures. To the best of our knowledge, 194 crowdworkers participated in the video recording for our dataset, making it the most subject-diverse RSL dataset and the second among all sign language datasets (see Table 1 for more details). The dataset was collected mainly indoors and varied in scenes and lighting conditions.

\begin{figure}
  \centering
  \includegraphics[width=\linewidth] {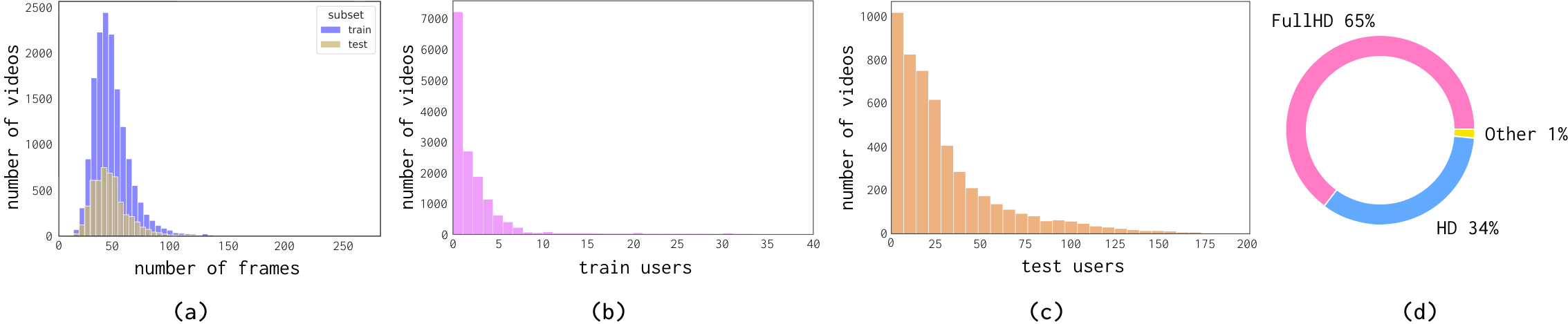}
  \caption{Video length, resolution and user's splitting analysis. (a) Videos' number of frames distribution divided into sets, (b) distribution of recorded video by users in train, and (c) test, (d) video resolution ratio.}
  \label{fig: stat}
\end{figure}

\textbf{Video Quality.} The videos were recorded primarily in HD and FullHD formats (see Figure~\ref{fig: stat}d). About 86\% of the videos are oriented vertically, 13\% are oriented horizontally, and 1\% are in square format. The number of frames distribution is also shown in Figure~\ref{fig: stat}a. The average video length is 1.67 seconds, and the overall duration of the dataset is about 19.81 hours. 

\textbf{Data Splitting.} The data was split into training (75\%) and test (25\%) sets, containing 15 and 5 video samples for each class, respectively. The numbers of subjects in training and test sets are equal 112 and 174, respectively. Note that groups of subjects in these two sets intersect; however, we tried to minimize it by filling out the test set with inactive signers (see Figure~\ref{fig: stat}b-c - the test set consists mainly of signers who have uploaded a small number of videos). This approach minimizes the intersection of signers in the training and test sets, reducing the risk of model overfitting.

\vspace{-0.8cm}
\section{Experiments}

\textbf{Models.} Addressing the challenge of recognizing sign language necessitates the utilization of formidable and lightweight models endowed with the capacity to analyze video data. Multiscale Vision Transformer (MViT)~\cite{fan2021multiscale} model was specifically designed for video recognition tasks and provides a significant performance gain over concurrent video transformers that rely on large-scale external pre-training and are several times more costly in computation and parameters. Creating a multiscale pyramid of features, MViT models effectively connect the principles of transformers with multiscale feature hierarchies. An Improved MViT architecture (MViTv2)~\cite{li2022mvitv2} proves to be a robust general backbone for computer vision tasks in the video domain. 
\begin{figure}
  \centering
  \includegraphics[width=\linewidth] {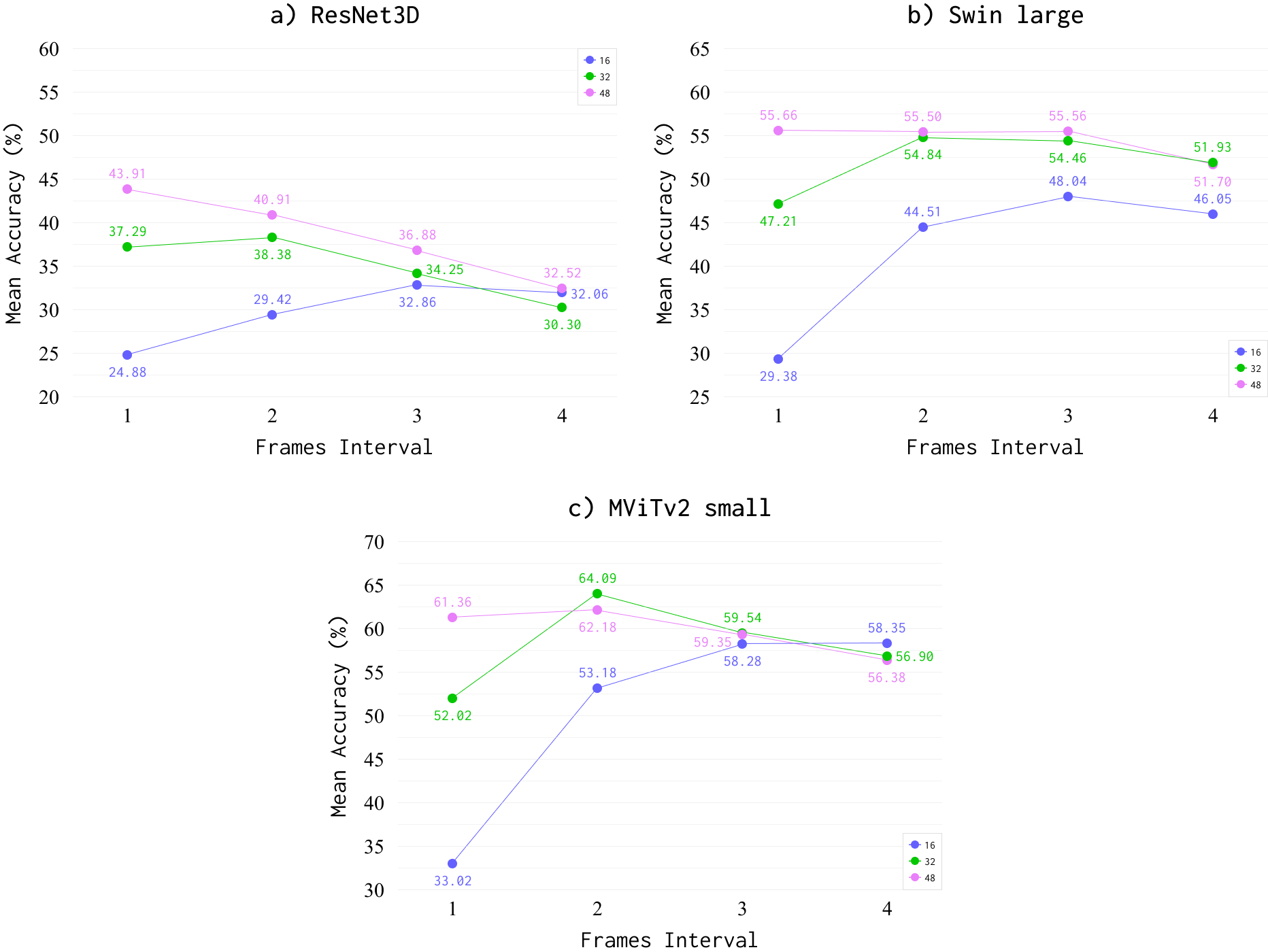}
  \caption{Mean accuracy is achieved by each model on the Slovo with different sampling strategies. Note that the graphs have various scales depending on the order of the metrics.}
  \label{fig: metrics}
\end{figure}

\vspace{-0.4cm}

The MViTv2 demonstrates state-of-the-art performance in various video recognition benchmarks and can accurately analyze video input. Therefore its small version was chosen as the baseline, we utilized Swin-large~\cite{liu2022video} and ResNet3D-50~\cite{resnet}, all pre-trained on the Kinetics dataset.

\textbf{Data Pre-processing.} The samples were resized on the maximum side to 300 pixels. MViTv2-small and Swin-large trained with a horizontal flip and sharpness augmentations, whereas ResNet3D-50 -- with the same horizontal flip, salt random noise, and color jitter. Horizontal flip augmentation is used to bring the data distribution to the real because RSL signs do not change the meaning of mirror reflection. Finally, the videos were padded to (300, 300) and randomly cropped to 224 pixels. 

\textbf{Implementation Details.} Several sampling strategies were tested with a different number of frames from [16, 32, 48] and a frame interval from 1 to 4. We also checked models trained on 64 frames, which generated poor results. Frame intervals are limited to 4 because skipping more frames in the SLR task can miss important information about the sign. We trained all 36 models over 120 epochs with a learning rate 0.001, AdamW~\cite{loshchilov2019decoupled} optimizer employing betas (0.9, 0.999), and weight decay 0.05. Two schedulers -- LinearLR and CosineAnnealingLR~\cite{loshchilov2017sgdr} -- were used to optimize the Swin-large and ResNet3D-50 training processes. Only LinearLR was used for MViTv2-small. The information about their parameters is in the repository.

\textbf{Results.} Since each gesture corresponds to 20 video samples, we validated the models on the test set. Figure \ref{fig: metrics} shows the results of each chosen model separately with a different number of frames and frame interval. We can observe that MViTv2-small, with 32 frames and an interval of 2, vastly outperforms other models due to its video purpose. We attribute the notably lower metrics of ResNet3D-50 to the superior performance of vision transformers compared to convolutional architectures in the domain of videos.





\section{Conclusion}
In this paper, we proposed the new Russian Sign Language dataset Slovo and a pipeline for creating diverse video data despite a specific domain. The Slovo is divided into 1,000 classes, each corresponding to 20 videos from 194 signers. It can favorably influence the development of sign language recognition in the Russian domain. Besides, several models were trained and evaluated on Slovo to demonstrate its teaching ability. Shortly, we plan to expand our dataset: increase the number of classes and the number of examples per class and collect not only words but also phrases, n-grams, and sentences. Furthermore, since SL is more expressive than regular ones: not only hand gestures are used to indicate words, but also facial expressions, articulation, and body posture, multimodal models can potentially improve SLR results. The current dataset, pre-trained models, and demo are publicly available in the repository.

\bibliographystyle{splncs04}
\bibliography{egbib}

\end{document}